\documentclass[10pt,twocolumn,letterpaper]{article}

\usepackage{iccv}
\usepackage{times}
\usepackage{epsfig}
\usepackage{graphicx}
\usepackage{amsmath}
\usepackage{amssymb}
\usepackage{enumitem}
\usepackage{paralist}

\usepackage{color}

\usepackage[pagebackref=true,breaklinks=true,letterpaper=true,colorlinks,bookmarks=false]{hyperref}

\iccvfinalcopy 

\def\iccvPaperID{130} 

\ificcvfinal\fi

\makeatletter
\long\def\comment#1{}
\long\def\@makecaption#1#2{
   \setbox\@tempboxa\hbox{\small \noindent {\small \bf #1.}~\small#2}
   \setlength{\@ctmp}{\hsize}
   \addtolength{\@ctmp}{-\@figindent}\addtolength{\@ctmp}{-\@figindent}
   \ifdim \wd\@tempboxa >\@ctmp
      {\small {\small\bf #1.}~\small#2\par}
   \else
      \hbox to\hsize{\hfil\box\@tempboxa\hfil}
  \fi\vspace*{\belowcaptionskip}}

\renewcommand\section{\@startsection {section}{1}{\z@}%
                                   {-.8ex \@plus -2ex \@minus -.2ex}%
                                   {.5ex \@plus.2ex}%
                                   {\normalfont\large\bfseries\raggedright}}
\renewcommand\subsection{\@startsection{subsection}{2}{\z@}%
                                   {-.6ex \@plus -2ex \@minus -.2ex}%
                                   {.5ex \@plus.2ex}%
{\normalfont\large\bfseries\raggedright}}
\renewcommand\subsubsection{\@startsection{subsubsection}{3}{\z@}%
                                     {-.5ex\@plus -.2ex \@minus -.2ex}%
                                     {.2ex \@plus .2ex}%
                                     {\normalfont\large\bfseries\raggedright}}

  \renewenvironment{thebibliography}[1]{%
    \begin{oldthebibliography}{#1}%
      \small
      \setlength{\parskip}{0pt}%
      \setlength{\itemsep}{0pt}%
  }%
  {%
    \end{oldthebibliography}%
  }

\renewcommand\normalsize{%
   \@setfontsize\normalsize\@xpt\@xiipt
   \abovedisplayskip 4\p@ \@plus2\p@ \@minus5\p@
   \abovedisplayshortskip \z@ \@plus3\p@
   \belowdisplayshortskip 2\p@ \@plus3\p@ \@minus3\p@
   \belowdisplayskip \abovedisplayskip
   \let\@listi\@listI}
\renewcommand\small{%
   \@setfontsize\small\@ixpt{11}%
   \abovedisplayskip 2.5\p@ \@plus3\p@ \@minus4\p@
   \abovedisplayshortskip \z@ \@plus2\p@
   \belowdisplayshortskip 2\p@ \@plus2\p@ \@minus2\p@
   \def\@listi{\leftmargin\leftmargini
               \topsep 4\p@ \@plus2\p@ \@minus2\p@
               \parsep 2\p@ \@plus\p@ \@minus\p@
               \itemsep \parsep}%
   \belowdisplayskip \abovedisplayskip
}

\def\tightmath{
\abovedisplayskip=3pt plus 2pt minus 1pt
\abovedisplayshortskip=0pt plus 1pt minus 1pt
\belowdisplayskip=3pt plus 2pt minus 1pt
\belowdisplayshortskip=0pt plus 1pt minus 1pt }
\def\crushmath{
\abovedisplayskip=1pt plus 1pt minus 2pt
\abovedisplayshortskip=1pt plus 1pt minus 2pt
\belowdisplayskip=1pt plus 1pt minus 2pt
\belowdisplayshortskip=1pt plus 1pt minus 2pt }
\tightmath
\crushmath

\makeatother

\begin{document}

\title{Learning Concept Embeddings with Combined Human-Machine Expertise}

\author{
Michael J. Wilber$^{1,2}$
$\qquad$
Iljung S. Kwak$^3$
$\qquad$
David Kriegman$^{3,4}$
$\qquad$
Serge Belongie$^{1,2}$\\
\\
$^1$ Department of Computer Science, Cornell University\quad$^2$ 	Cornell Tech\\
$^3$ Department of Computer Science and Engineering, UC San Diego\quad$^4$ Dropbox
}

\maketitle
\thispagestyle{empty}

\begin{abstract}
This paper presents our work on ``SNaCK,'' a low-dimensional concept embedding algorithm that combines human expertise with automatic machine similarity kernels. Both parts are complimentary: human insight can capture relationships that are not apparent from the object's visual similarity and the machine can help relieve the human from having to exhaustively specify many constraints. We show that our SNaCK embeddings are useful in several tasks: distinguishing prime and nonprime numbers on MNIST, discovering labeling mistakes in the Caltech UCSD Birds (CUB) dataset with the help of deep-learned features, creating training datasets for bird classifiers, capturing  subjective human taste on a new dataset of 10,000 foods, and qualitatively exploring an unstructured set of pictographic characters. Comparisons with the state-of-the-art in these tasks show that SNaCK produces better concept embeddings that require less human supervision than the leading methods.



\end{abstract}

\section{Introduction}\label{sec:intro}
Supervised learning tasks form the backbone of many state-of-the-art computer vision applications. They help researchers classify, localize, and characterize actions and objects.
However, if the researcher's goal is instead to interactively explore the latent structure of a dataset, discover novel categories, or find labeling mistakes, it is unclear what kind of supervision to use. Sometimes the data does not fall into well-defined taxonomic categories, or perhaps it is simply too expensive to collect labels for every object. Sometimes the expert wishes to capture a \emph{concept}---some intuitive constraint that they cannot articulate---about how the data should be structured, but does not have the time to specify this concept formally. If we wish to build models that capture concepts, we need a new approach.




\begin{figure}[t]
  \includegraphics[width=\linewidth]{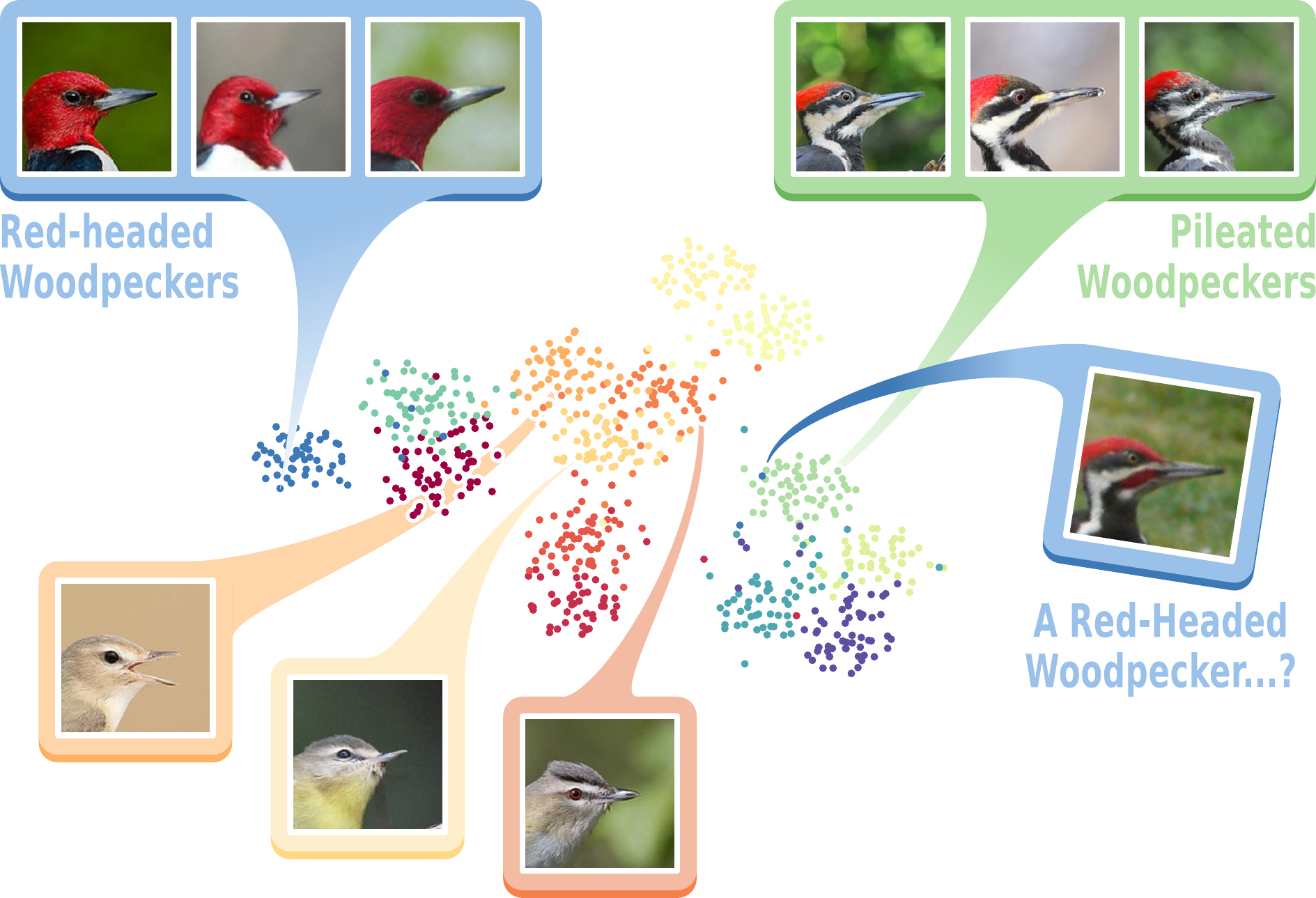}
  \caption{Our SNaCK embeddings capture human expertise with the help of machine similarity kernels. For example, an expert can use this concept embedding of a subset of CUB-200 to quickly find labeling mistakes. Red-headed Woodpeckers are visually dissimilar to Pileated Woodpeckers, but SNaCK moved a Red-headed Woodpecker into the Pileated Woodpecker cluster because of its appearance. \textbf{This is probably a labeling mistake in CUB-200, and this SNaCK embedding helped us discover it.} The cluster of three visually similar vireo species in the embedding center may be another good place to look for label problems.
}
  \label{fig:snackpower}
\end{figure}

Our overall goal is to generate a \emph{concept embedding}. Distances within this space should correspond with a human's intuitive idea of how similar two objects are. Many researchers use similar embeddings to enhance the performance of classifiers~\cite{schroff_facenet:_2015,branson_ignorant_2014,wah_similarity_2014}, build retrieval systems~\cite{van_der_maaten_stochastic_2012,mcfee}, and create visualizations that help experts better understand high-dimensional spaces~\cite{demiralp_learning_2014,demiralp_visual_2014}.

\textbf{Concepts cannot always be inferred from appearance}. Within the past few years, huge research advances have begun to produce systems that are excellent at comparing images based on visual cues. For example, one can imagine building a CNN to compare food dishes based solely on their appearance. However, if the concept we wish to capture is similarity in taste, the task becomes harder. Although taste and appearance are often correlated, any poor diner who has confused guacamole and wasabi knows that foods that taste very different may look deceptively similar because the strongest visual cues may not be reliable. This particular taste difference is difficult to capture without expert guidance. Similarly, when classifying birds, the goal is often not to group similar-looking birds together, but to group birds of the same species together. Experts know that appearance is important for this classification task, but there are often large visual differences between the appearance of male and female birds of the same species or between juveniles and adults. In these cases, domain-specific expertise can greatly improve the resulting embedding.

\textbf{Expert annotations can be expensive to collect.} In order to capture abstract concepts known only by humans, the expert must provide \emph{hints}~\cite{abu-mostafa_machines_1995} 
to help guide the learning process. Unfortunately, asking experts to exhaustively and authoritatively annotate the dataset is not always possible~\cite{beijbom2012automated}. Further, hints are most useful when they are task-specific~\cite{demiralp_visual_2014}: if the user wishes to discover some relationship that is not apparent between objects, they should be able to specify whatever hints they feel would best capture those constraints. Previous work that uses perceptual annotations~\cite{van_der_maaten_stochastic_2012,cost-effective-hits} note that collecting all hints based on relative similarity comparisons can take quadratic or cubic cost. Hiring actual domain experts is often out of the question, and even crowdsourcing websites such as Mechanical Turk can be prohibitively expensive.

It seems reasonable that one can use machine kernels to speed up the process of collecting hints. 
In this work, we show how to overcome the inherent human scalability problems by using human hints to refine  a concept embedding generated by an automatic similarity kernel.
Our main contributions are as follows:
\begin{compactitem}
\item We present a \textbf{novel algorithm}, ``\emph{SNE-and-Crowd-Kernel Embedding}'' (SNaCK), that combines expert triplet hints with machine assistance to efficiently generate concept embeddings; 
\item We show how to use our SNaCK embeddings for tasks such as visualization, concept labeling, and perceptual organization, and show that SNaCK embeddings are competitive with the state of the art in these tasks.
\end{compactitem}
We also present the following minor contributions:
\begin{compactitem}
  \item A dataset of 950,000 crowdsourced perceptual similarity annotations on 10,000 food dishes from Yummly;
  \item A deep-learned food classifier that greatly improves upon the previous state-of-the-art performance on the Food-101 dataset~\cite{bossard_food-101mining_2014};
  \item A proof that two perceptual embedding algorithms in common use, CKL and t-STE, 
  are equivalent for the common 2D case for certain parameter settings. To our knowledge, this connection has never been acknowledged or explored before.
\end{compactitem}

\begin{figure}[t]
  \includegraphics[width=\linewidth]{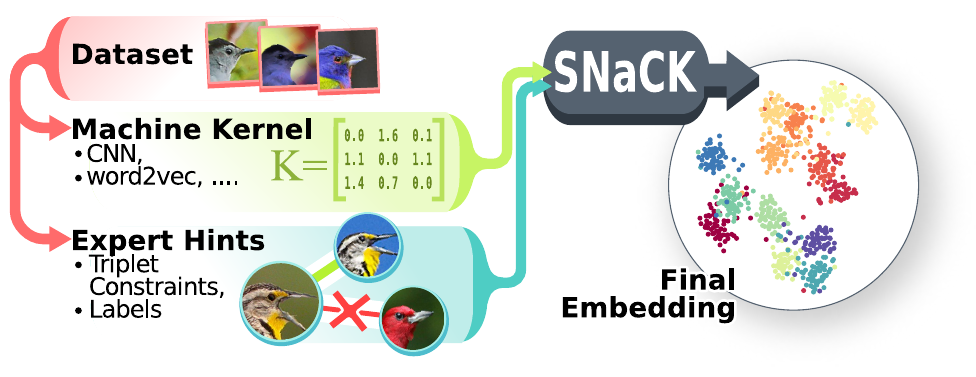}
  \caption{Overview of our SNE-and-Crowd-Kernel (``SNaCK'') embedding method. As input, SNaCK accepts a dataset of objects, a similarity kernel $K$, and a set of expert constraints in the form of ``Object $i$ should be closer to $j$ than it is to $k$'', which may be inferred from crowdsourcing or label information. The output is a low-dimensional concept embedding that satisfies the expert hints while preserving the structure of $K$.\label{fig:overview}}
\end{figure}

\section{Background and related work}

\textbf{Perceptual embeddings.} Our work builds upon a large body of existing perceptual embedding literature. Notably, our method combines aspects of both \emph{t-Distributed Stochastic Neighbor Embedding} (t-SNE, from \cite{van_der_maaten_visualizing_2008}) and \emph{Stochastic Triplet Embedding} (t-STE, from \cite{van_der_maaten_stochastic_2012}). The objective and motivation behind these approaches are fundamentally different: t-SNE creates a low-dimensional visualization using an automatic kernel from a higher-dimensional space and t-STE generates an embedding from scratch that satisfies as many human-provided similarity constraints as possible. Nevertheless, we show that they are complementary.
%
Interestingly, there is a strong mathematical similarity between t-STE and the \emph{Crowd Kernel Learning} (CKL) method described in \cite{tamuz_adaptively_2011}; in fact, in the supplementary material, we show that \emph{CKL and t-STE are equivalent} for certain parameter choices. To our knowledge, this connection has not been explored before.

\textbf{Triplet constraints and other kinds of hints}.
In our work, we use \emph{triplet constraints}, where the crowd or the expert provides tuples of the form $(i, j, k)$ to indicate that object $i$ seems more similar to object $j$ than $i$ does to object $k$. We take these constraints to mean that object $i$ should thus be \emph{closer} to object $j$ than $i$ is to $k$ in the desired concept embedding. These relative comparisons allow the expert to directly specify perceptual constraints about objects. When compared to other forms of supervision, triplets are one of the most flexible options in practical use because they do not rely on \emph{a priori} knowledge, are invariant to scale, and are stable between and within subjects. Consider other forms of supervision: placing objects into \emph{category labels} may not map to the abstract concepts the expert wishes to capture and it requires the entire taxonomy to be known up-front. Even with unlimited time and a patient expert, the label results may be subject to scrutiny: one human expert solving the ImageNet Large Scale Visual Recognition Challenge~\cite{DBLP:journals/corr/RussakovskyDSKSMHKKBBF14} took approximately a minute to label each image and still made 5.1\% error. The CUB-200~\cite{WahCUB_200_2011} dataset also has labeling errors, which we will show in Sec.~\ref{sec:birdlets-discovery}.

\emph{Pairwise similarity judgments} are another common form of supervision, but they have own problems. The classic 7-point Likert scale induces quantization into the metric and may not be reliable between people. Several researchers~\cite{miller1956magical, kendall1948rank, demiralp_learning_2014} note that methods based on triplet comparisons are more stable than such pairwise measures. In an experiment comparing the speed and effectiveness of pairwise, triplet, and spatial arrangement embeddings,~\cite{demiralp_learning_2014} found that triplet comparisons yield the least variance of human perceptual similarity judgments than other methods, though triplet tasks also took humans the longest to complete.
%
One disadvantage of triplet constraints is that triplet embeddings require at least $O(n^3)$ triplet constraints to be uniquely specified~\cite{kleindessner_uniqueness_2014}, even though many triplets are strongly correlated and do not contribute much to the overall structure~\cite{schroff_facenet:_2015}. This is why we propose using a machine vision system to do most of the heavy lifting and reduce the number of required triplet constraints.




\textbf{Incorporating human judgments in automatic systems.}
Of course, we are not the first researchers to show the benefits of combining human and machine expertise. For example,~\cite{branson_ignorant_2014,wah_similarity_2014} build a classification system by bringing humans ``into the loop'' at runtime. Other work allows humans to specify an attribute relationship to influence the label training~\cite{biswas_simultaneous_2013}. These approaches are most useful when classification is the end goal rather than visualization or perceptual organization.
%
Another branch of work starts from an automatically-generated distance matrix and uses human constraints to further refine the recovered clustering or distance metric, typically by asking the human to provide pairwise ``Must-link'' or ``Must-not-link'' constraints~\cite{lu_constrained_2008,xing_distance_2002,yi_semi-crowdsourced_2012,tang_enhancing_2007,yi_semi-supervised_2013}. In some works, the human can provide an attribute explanation for their choice~\cite{lad_interactively_2014}. 
In Sec.~\ref{sec:birdlets-discovery}, we show that our approach is competitive with many of these constrained clustering algorithms in a semi-supervised labeling task.

Other particularly relevant contributions re-cast t-STE as a multiple metric learning problem~\cite{zhang_jointly_2015}. Here, the humans are asked to evaluate multiple aspects of objects' similarity (eg. similarity of different parts), and the final embedding is learned to jointly satisfy as many aspects as possible. Similarly,~\cite{amid_multiview_2015} learns multiple maps from a single set of triplet questions. Our work is similar in spirit, but our focus on jointly learning both human and machine-judged similarity rather than just multiple aspects of human similarity
%
sets us apart from these works and others such as~\cite{gomes_crowdclustering_2011}, which focus on creating more efficient user interfaces to gather data from crowdsourcing without using machine vision to accelerate the process.


\textbf{Embeddings from deep learning and Siamese networks.} Finally, an interesting branch of work revolves around teaching CNNs to satisfy triplet questions as part of the overall pipeline~\cite{wang_learning_2014,wu_online_2013}. One method based on this approach currently holds the state-of-the-art accuracy on the LFW face verification challenge~\cite{schroff_facenet:_2015}. Methods like this are very appealing if one wishes to build a classifier. Other methods~\cite{hadsell_dimensionality_2006} train Siamese networks on pairwise distance matrices to output the embedding directly. Though our work does use deep learning as part of our pipeline, deep learning is not necessary for our approach.







\section{``SNE-and-Crowd-Kernel'' (SNaCK) embeddings}
Our hybrid embedding algorithm, \emph{SNE-and-Crowd-Kernel} (SNaCK), jointly optimizes the objective functions of two different low-dimensional embedding algorithms.\footnote{Our code is available on the companion website, \url{http://vision.cornell.edu/se3/projects/concept-embeddings}} The first algorithm, t-SNE (\emph{t-Distributed Stochastic Neighbor Embedding}~\cite{van_der_maaten_visualizing_2008}), uses a distance matrix to construct a low-dimensional embedding. Its goal is to ensure that objects which are close in the original high-dimensional space are also close in the low-dimensional output without constraining points that are far in the original space. The second method, t-STE \emph{(Stochastic Triplet Embedding}~\cite{van_der_maaten_stochastic_2012}), allows experts to supply triplet constraints that draw from their domain knowledge and task-specific hints. We will show that this surprisingly simple joint optimization can capture the benefits of both objectives. See Fig.~\ref{fig:overview} for an overview.

\subsection{Formulation}
Consider $N$ objects. We wish to produce a $d$-dimensional embedding $Y \in \mathcal{R}^{N \times d}$. Let $K \in \mathcal{R}^{N \times N}$ be a distance matrix, and let $T = \{t_1, \ldots, t_M\}$ be a set of triplet constraints. Each constraint $t_\ell = (i,j,k)$ implies that in the final embedding, object $i$ should be closer to object $j$ than it is to $k$, meaning $\|y_i - y_j\|^2 \leq \|y_i - y_k\|^2$. According to~\cite{van_der_maaten_visualizing_2008}, the loss function for t-SNE can be interpreted as finding the low-dimensional distribution of points that maximizes the information gain from the original high-dimensional space.
\begin{equation}
  C_{tSNE} = \sum_{i \neq j} p_{ij} \log \frac{p_{ij}}{q_{ij}},
\end{equation}
where
\begin{align}
  p_{j|i} &= \frac{\exp( - K_{ij}^2 / 2\sigma_i^2)}
{\sum_{k \neq i} \exp( - K_{ik}^2 / 2\sigma_i^2)}\\
  p_{ij} &= \frac{1}{2N} (p_{j|i} + p_{i|j})\\
  q_{ij} &= \frac{(1 + \|y_i-y_j\|^2)^{-1}}
{\sum_{k \neq l} (1+\|y_k-y_l\|^2)^{-1}}
\end{align}
and $\sigma_i$ is chosen to satisfy certain perplexity constraints.

The loss function for t-STE, given in~\cite{van_der_maaten_stochastic_2012}, can be interpreted as the joint probability of independently satisfying all triplet constraints. It is defined as
\begin{equation}
C_{tSTE} = \sum_{(i,j,k) \in T} \log p^{tSTE}_{(i,j,k)},
\end{equation}
where
\begin{equation}
p^{tSTE}_{(i,j,k)} = \frac{
\left( 1 + \frac{\|y_i - y_j\|^2}{\alpha} \right)^{-\frac{1+\alpha}{2}}
}{
\left( 1 + \frac{\|y_i - y_j\|^2}{\alpha} \right)^{-\frac{1+\alpha}{2}}
+
\left( 1 + \frac{\|y_i - y_k\|^2}{\alpha} \right)^{-\frac{1+\alpha}{2}}
}
\end{equation}
Interestingly, when $\alpha=1$ (as suggested in~\cite{van_der_maaten_stochastic_2012} for two-dimensional visualizations), $C_{tSTE}$ becomes a special case of the cost function $C_{CKL}$ from~\cite{tamuz_adaptively_2011} for certain parameter choices. We explore this relationship in the supplementary material. Because they are equivalent, we use $C_{tSTE}$ in our cost function, defined as
\begin{equation}
C_{SNaCK} = \lambda \cdot C_{tSTE} \;+\; (1 - \lambda) \cdot C_{tSNE}
\end{equation}
To optimize this cost, we use gradient descent on $\frac{\partial C_{SNaCK}}{\partial Y}$. Our implementation derives from the t-SNE implementation in \verb+scikit-learn+, so we inherit their optimization strategy. In particular, we use t-SNE's early exaggeration~\cite{van_der_maaten_visualizing_2008} heuristic for 100 iterations and then continue until the 300th iteration.

The $\lambda$ parameter specifies the relative contribution of the machine-computed kernel and the human-provided triplet constraints on the final embedding. For each experiment, we pick $\lambda$ up front such that the norm of $\frac{\delta  C_{tSTE}}{\delta Y}$ is approximately equal to $\frac{\delta C_{tSNE}}{\delta Y}$ in cross validation.

\subsection{SNaCK example: MNIST}\label{sec:mnist}
\begin{figure}[tdp]
  \includegraphics[width=\linewidth]{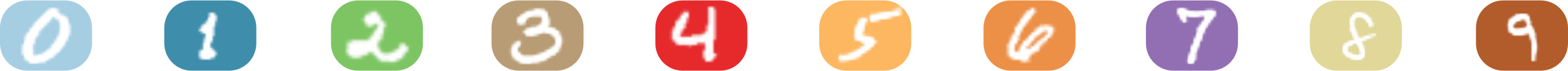}
  \includegraphics[width=\linewidth]{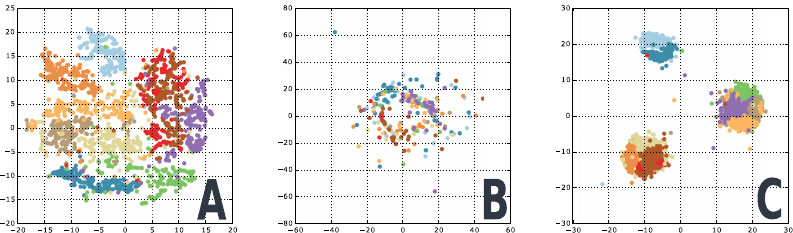}
  \caption{A simple MNIST example to illustrate the advantages of SNaCK's formulation. Suppose an expert wishes to group MNIST by some property that is not visually apparent, in this case: prime, composite or $\{0,1\}$. (A) shows t-SNE on 2,000 MNIST digits using flattened pixel intensities. (B) shows t-STE on 1,000 triplets of the form $(i,j,k)$, where $i$ and $j$ share the same concept but $k$ does not. (C) shows a SNaCK embedding using the same flattened pixel intensities and the same triplet constraints. The SNaCK embedding is the only one that captures the intra-class structure from (A) and the desired abstract grouping of (B). See~\ref{sec:mnist} for details.}
  \label{fig:mnist}
\end{figure}

To briefly illustrate why this formulation is better than t-STE or t-SNE alone, Fig.~\ref{fig:mnist} shows a toy example on MNIST data. In this example, suppose the expert wishes to capture the concept of primality by partitioning the dataset into prime numbers $\{2,3,5,7\}$, composite numbers $\{4,6,8,9\}$ and $\{0,1\}$. Also, for the purpose of this simple example, assume that rather than labeling the digits directly, the expert compares images based on concept similarity, \emph{i.e.}, primes are more similar to primes than to other images. By running t-SNE on flattened pixel intensities, Fig.~\ref{fig:mnist}~(A) illustrates that the embedding does a reasonable job of clustering numbers by their label but clearly cannot understand primality because this concept is not apparent from visual appearance. To compensate, we sample triplet constraints of the form $(i,j,k)$ where $i$ and $j$ share the same concept and $k$ does not. However, we only sample 1,000 constraints for these 2,000 images. t-STE (B) attempts to discover the differences between the numbers in a ``blind'' fashion, but since it cannot take advantage of any visual cues, the underconstrained points are effectively random. If given many more constraints, eventually t-STE can only collapse everything into three points for each of the three abstract concepts. Our SNaCK embedding (C) displays the desired high-level concept grouping into primes/non-primes/others, and it can capture the structure of each class. Points with too few constraints are corrected by the t-SNE loss and the t-STE loss captures the appropriate structure.

\section{Experiments}
Our MNIST example demonstrates SNaCK's utility in a domain where concepts can be derived from category labels and everything is known \emph{a priori}. How does SNaCK perform on domains where a fixed taxonomy or fixed category labels are not necessarily known up front? To explore this question, we perform a series of experiments: first, we showcase SNaCK's ability to help label a subset of CUB-200 in a semi-supervised fashion. In this setting, SNaCK learns concepts that are equivalent to category labels and outperforms other semi-supervised learning algorithms.
Second, our experiments on a dataset of 10,000 unlabeled food images demonstrate SNaCK's ability to capture the concept of food taste using crowdsourcing. We evaluate the embedding's generalization error on a held-out set of crowdsourced triplet constraints.
Finally, we showcase SNaCK's ability to embed a set of pictographic characters, showing how an expert can interactively explore and refine the structure of an embedding where no prior knowledge is available.



\subsection{Incrementally labeling CUB-200-2011}\label{sec:birdlets-discovery}

\begin{figure}[h]
\includegraphics[width=1\linewidth]{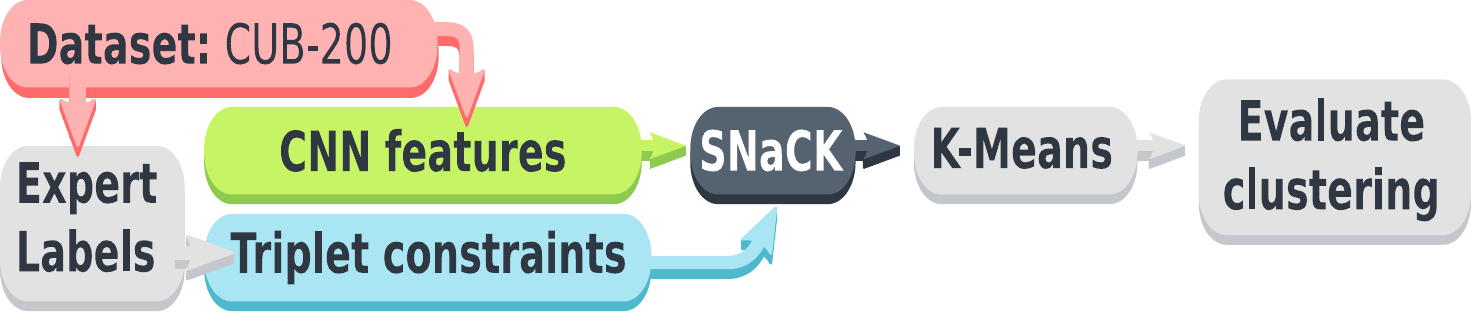}
\caption{Experiment overview on CUB-200. See text for details.\label{fig:flowchart-birds}}
\end{figure}

In this scenario, we show how SNaCK embeddings can help experts label a new dataset. Suppose an expert has a large dataset with category annotations and an unlabeled smaller set containing new classes similar to those they already know. The expert wishes to use their extensive preexisting knowledge to quickly label the new set with a minimum amount of human effort. Our goal is to show that SNaCK allows the expert to collect high-quality labels more quickly than other methods. Here, the ``concepts'' we learn are equivalent to category labels. These experiments are inspired by~\cite{lee_learning_2011}. See Fig.~\ref{fig:flowchart-birds}.

\textbf{Dataset.} For this task, we use the ``Caltech-UCSD Birds 200-2011'' (CUB-200) dataset~\cite{WahCUB_200_2011}. We assume the expert has access to all images and labels of 186 classes in the dataset (to train a machine kernel) and wishes to quickly label a testing set of 14 classes of woodpeckers and vireos. This subset contains 776 images and was defined in~\cite{farrell_birdlets:_2011}. We only use profile-view bird images where a single eye and the beak is visible. Images are rotated, scaled, and possibly flipped so the eye is on the left side of the image and the beak is on the right side; part locations are collected using crowdsourcing. The image is then cropped to the head. This is the same normalization strategy as~\cite{BransonVBP14}.

\textbf{Automatic similarity kernel.} To generate $K$, we fine-tune a CNN to a classification task on all images in the 186 known classes. This allows the expert to leverage their extensive pre-existing dataset to speed up label collection for the novel classes. Our network is a variation of the ``Network-in-Network'' model~\cite{lin_network_2013}, which takes cropped normalized bird heads as input and outputs a 186-dimensional classification result. We started from the pre-trained ImageNet model in the Caffe model zoo~\cite{jia2014caffe} and fine-tuned the network for 20,000 iterations on an Amazon EC2 GPU instance. To do this, we replaced the last layer with a 186-class output and reduced the learning rate for the other layers to a tenth of the previous value.\footnote{When trained using the standard training/testing protocol on all of CUB-200, this kind of model achieves 74.91\% classification accuracy, which is comparable to the state-of-the-art~\cite{BransonVBP14}.} Finally, $K^{CNN}_{i,j}$ is the Euclidean distance between features in the final layer before softmax. To evaluate the importance of specialized kernels, we also compare this $K^{CNN}$ kernel to Euclidean distances between pre-trained GoogLeNet~\cite{googlenet} features, and Euclidean distance between HOG features.

\textbf{Expert constraints.} To generate triplet constraints in a semi-supervised fashion, we reveal the labels for $n$ images of the dataset and sample all triplets between these images that satisfy same/different label constraints to generate $T_n = \{(i, j, k) \;|\; \ell_i = \ell_j \neq \ell_k, \max(i,j,k)\leq n\}$. This allows us to vary the amount of expert effort required to label the novel images. Note that in this test, our concepts to learn are equivalent to class labels, so all of our sampled constraints are derived from ground truth. Our food experiments, described in the next section, will demonstrate SNaCK's ability to learn more abstract concepts captured from subjective human judgments.

\begin{figure}[tdp]
  \includegraphics[width=\linewidth]{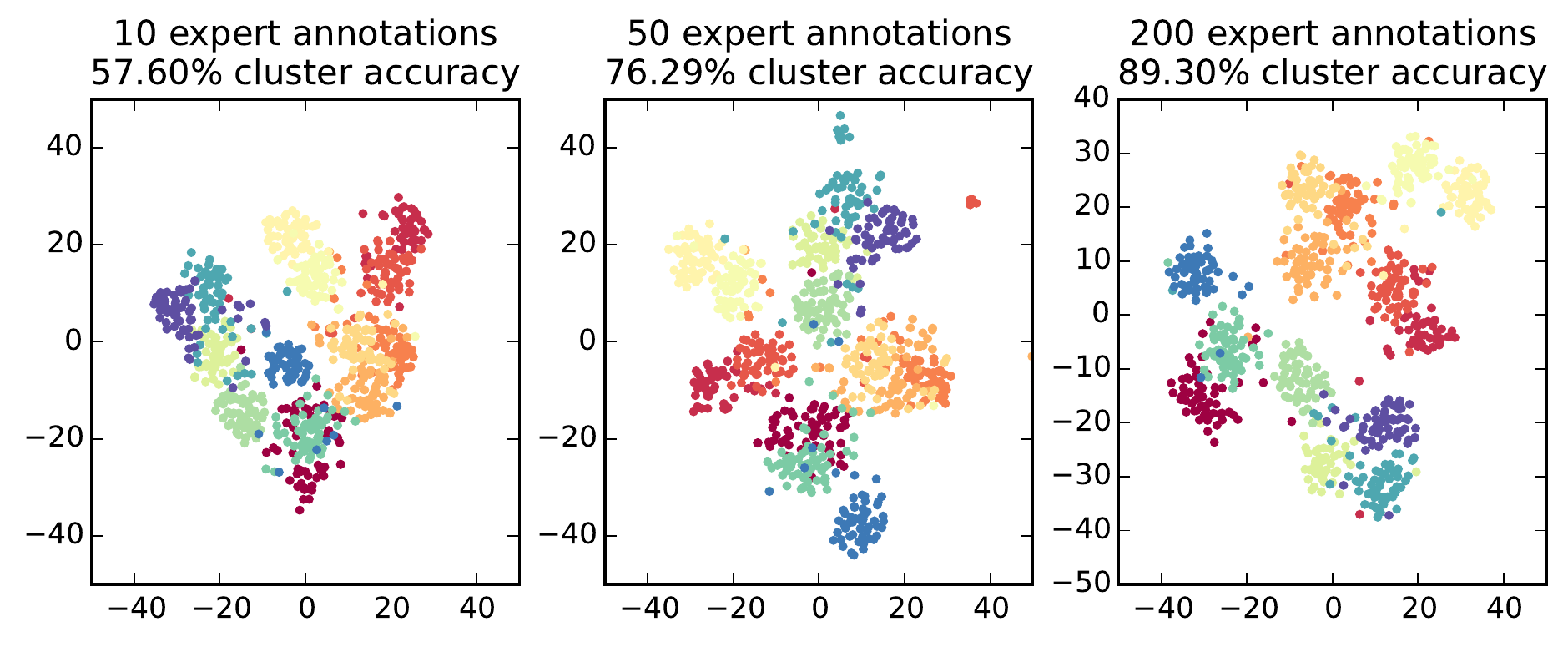}
  \caption{Embedding examples on CUB-200 Woodpeckers and Vireos, showing the ``SNaCK'' method with (left-to-right) 10, 50, and 200 expert label annotations. Colors indicate ground truth labels. As the number of expert annotations increases, clusters within the SNaCK embedding become more consistent. \label{fig:birdlets-examples}}
\end{figure}

\textbf{Comparisons and metrics.} To perform labeling with SNaCK, we generate an embedding of all 776 images and use KMeans to find clusters. To evaluate, we assign all points within each discovered cluster to their most common ground truth label and calculate the accuracy of this assignment. See Fig.~\ref{fig:birdlets-examples} for example embeddings varying the number of expert label annotations. We compare against other semi-supervised learning and constrained clustering systems: \emph{Label Propagation}~\cite{label-prop}, the multiclass version of the \emph{Constrained Spectral Clustering KMeans} (CSPKmeans) method described in~\cite{wang_constrained_2012}, and \emph{Metric Pairwise-Constrained KMeans} (MPCKmeans)~\cite{bilenko_integrating_2004}. Label propagation uses $K^{CNN}$ and the $n$ revealed labels. The constrained clustering systems use $K^{CNN}$ and pairwise ``Must-Link'' and ``Cannot-Link'' constraints as input, so we reveal $n$ image labels and sample all possible pairwise constraints between them. As baselines, we calculate CNN features and try to cluster them with KMeans and spectral clustering, which do not benefit from extra human effort. Finally, we also compare against the cluster results of using K-Means on a t-STE embedding from the same triplet constraints used by SNaCK.

\begin{figure}[tdp]
  \includegraphics[width=\linewidth]{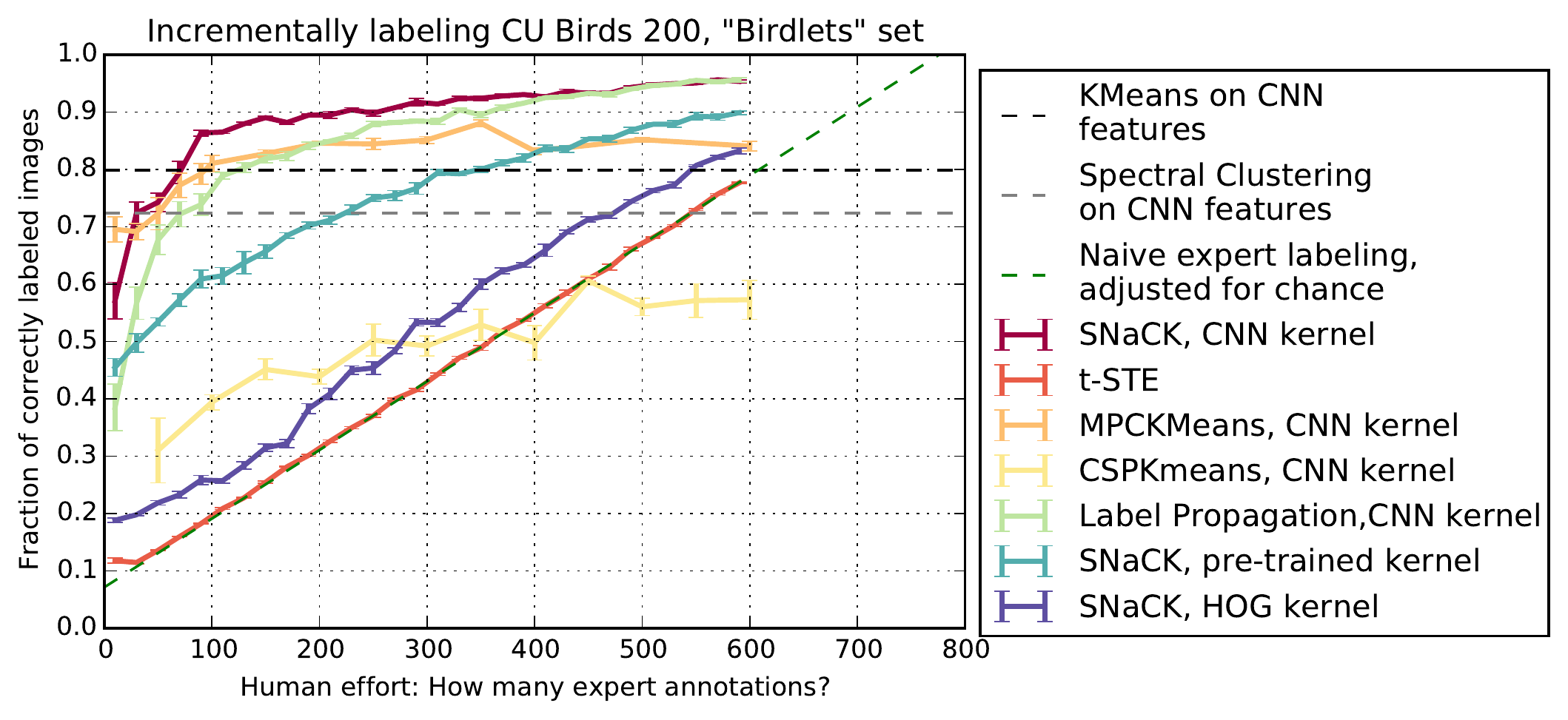}
  \caption{Incremental labeling accuracy of several semi-supervised methods. X axis: how many labels are revealed to each algorithm. Y axis: Dataset labeling accuracy. Error bars show standard error of the mean ($\sigma/\sqrt{n}$) across five runs. With 14 clusters, chance is $\approx 0.071$. See Sec.~\ref{sec:birdlets-discovery} for details. \label{fig:birdlets-discovery}}
\end{figure}

\textbf{Results} are shown in Fig.~\ref{fig:birdlets-discovery}. 
SNaCK outperforms all other algorithms, but label propagation and MPCKMeans also perform well. CSPKmeans is eventually outpaced by naively asking the expert for image labels, perhaps because it was designed for the two-class setting rather than our 14-class case. These experiments show that t-STE benefits from an automatic machine kernel (compare SNaCK to t-STE), but we can improve the machine kernel with a small number of expert annotations (compare KMeans or Spectral Clustering to SNaCK).

Using a kernel that captures bird similarity well is particularly important for this task. All of the algorithms which use $K^{CNN}$ generally outperform SNaCK when using a pre-trained GoogLeNet kernel. HOG features, which use no learning, are only slightly better than naive labeling. Finally, t-STE cannot use any visual kernel, so it can only consider the images the expert already revealed.

Sometimes the machine kernel disagrees with the expert hints. This may happen for interesting reasons, such as mistakes in the training data. For example, Figure~\ref{fig:snackpower} shows an instance of a Red-headed Woodpecker that was moved into a cluster containing many Pileated Woodpeckers. Even though the human constraints encourage this sample to lie near similarly labeled examples, this individual looks overwhelmingly similar to a Pileated Woodpecker, so the t-SNE loss overpowered the t-STE constraints. If the embedding is colored with ground truth labels, this mistake shows up as a single differently-colored point in the expected cluster, which is immediately apparent to an expert.

\subsubsection{Discovering labels for semi-supervised classifiers}\label{sec:birdlets-classification}
Does better incremental labeling translate into increased classification performance? In this scenario, we extend our previous experiment: we use SNaCK to discover labels for a training set and measure the accuracy of a simple SVM classifier on a testing set. Our goal is to decide whether just letting an expert reveal $n$ labels and training on this smaller set is better than revealing $n$ labels and using SNaCK to discover the rest. Will a classifier trained on many noisy, discovered labels perform differently than a classifier trained on a smaller, perfect training set?


\textbf{Dataset.} This task uses the same set of 14 woodpeckers and vireos from CUB-200 as before, but the procedure is different. We split our set into 396 training and 380 testing images using the same train/test split as CUB-200. We then discover labels on the training images using varying numbers of expert annotations and train a linear SVM classifier on all CNN features using the discovered labels. Finally, we report accuracy on the 380 testing images. The idea is that the quality of the discovered labels influences the accuracy of the classifier: a poor labeling method will cause the classifier to be trained on incorrect labels. Because all methods use the same type of classifier, we are evaluating the quality of our \emph{discovered labels}, not the classifier itself.

\textbf{Comparisons.} As a baseline, we compare SVM classifiers trained on SNaCK-discovered labels to an SVM classifier trained on a smaller, better set of $n$ correct labels provided by expert ground truth. This corresponds to the ``Naive Human Sampling'' method in Fig.~\ref{fig:birdlets-discovery}.
We also compare baselines where the SVM training set labels are discovered using KMeans, spectral clustering, and label propagation.

\begin{figure}[tdp]
  \includegraphics[width=\linewidth]{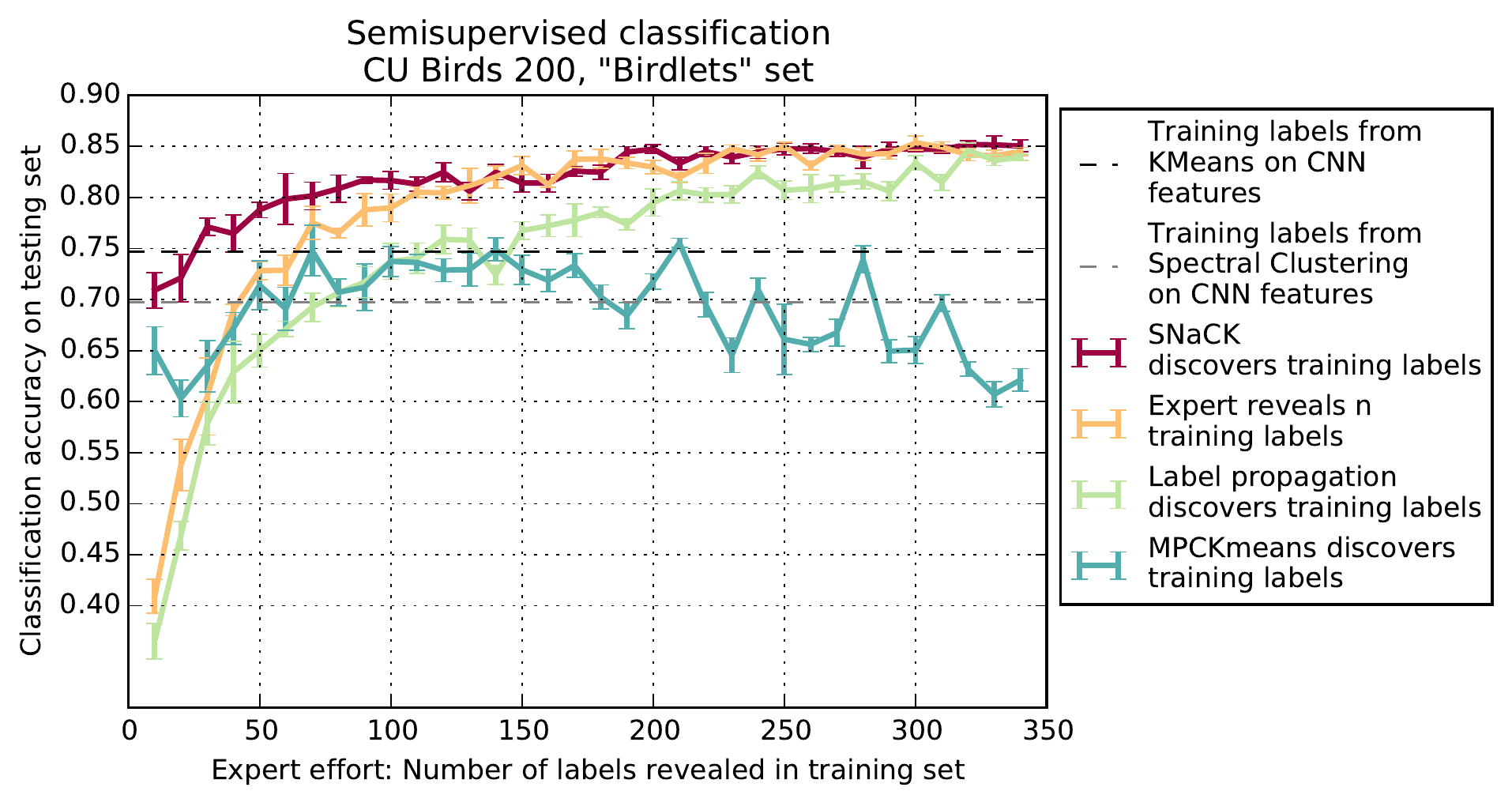}
  \caption{Classification accuracy of a linear SVM classifier trained on labels discovered by different methods. X axis: how many training labels are revealed to each algorithm. Y axis: Accuracy of classifier trained with these labels on the test set. Error bars show standard error of the mean ($\sigma/\sqrt{n}$) across five runs. See Sec.~\ref{sec:birdlets-discovery} for details. \label{fig:birdlets-classification}}
\end{figure}

\textbf{Results} are shown in Fig.~\ref{fig:birdlets-classification}. Classifiers trained on noisy labels discovered from SNaCK embeddings significantly outperform classifiers that are trained on smaller training sets, even though many of SNaCK's labels are incorrect. This is particularly true for fewer than 50 annotations. Accuracy of SNaCK, Label Propagation, and naive label sampling saturates at about 85\%, which is likely due to the linear SVM's limited generalization ability.

Interestingly, classification accuracy of labels discovered with MPCKMeans does not monotonically improve with more expert annotations. This surprises us, but Fig.~\ref{fig:birdlets-discovery} does show that MPCKmeans saturates to a smaller value in our semi-supervised labeling experiments, indicating that it cannot perfectly satisfy (and thus does not benefit from) additional constraints.

Using SNaCK, an expert can build a classifier that achieves 78.8\% classification accuracy by labeling 50 images (12.6\% of the dataset). A standard SVM that achieves this level of accuracy requires a training set of 95 perfectly labeled images, showing that SNaCK can cut down the expert's work load to build training sets for classifiers.

\subsection{Experiments on Yummly-10k}\label{sec:experiments_yummly10k}
\begin{figure*}[tdp]
  \includegraphics[width=\linewidth]{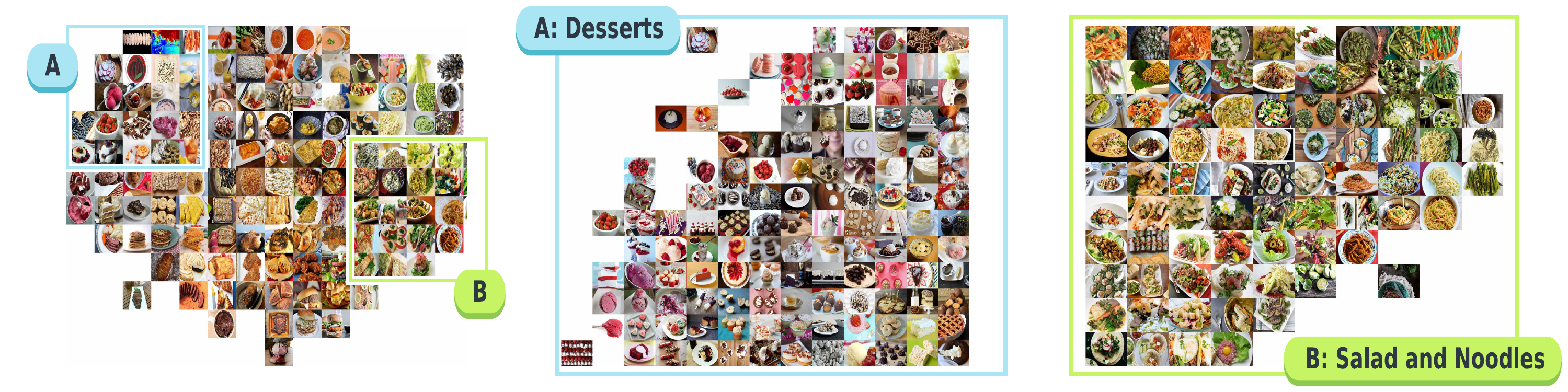}
  \caption{Left: Example SNaCK embedding on Yummly-10k, combining expertise from Kernel 2 (CNN features) and 950,000 crowdsourced triplet constraints. Middle/right: Close-ups of the embedding. On a large scale, SNaCK groups major food kinds together, such as desserts, salad, and main courses. On a small scale, each food closely resembles the taste of its neighbors. See the supplementary material version for larger versions of this figure.\label{fig:food-embeddings}}
\end{figure*}
\begin{figure}[h]\centering
\includegraphics[width=0.8\linewidth]{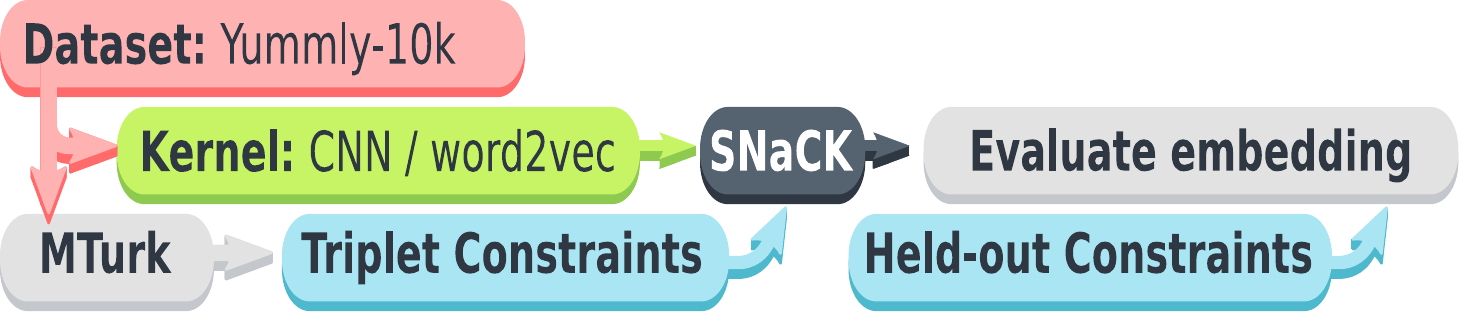}
\caption{Experiment Overview for Yummly-10k. See text for details.\label{fig:flowchart-yummly10k}}
\end{figure}

In this scenario, we use SNaCK to generate embeddings of food dishes. 
The goal is to create a concept embedding that captures the concept of taste. Two foods should be close in this embedding if they \emph{taste similar}, according to subjective human judgments. This is different from the earlier bird experiments because we can no longer rely on labels or taxonomies to help refine the embedding; all expert hints must come directly from unquantified human perception annotations. See Fig.~\ref{fig:flowchart-yummly10k}.

\textbf{Dataset.} For this experiment, we used 10,000 food images from the \emph{Yummly} recipe web site, dubbed \emph{Yummly-10k}. This data contains a variety of meals, appetizers, and snacks from different cultures and styles. We filtered the images by removing all images shorter or thinner than 300 pixels and removed all drinks and non-edibles. As metadata, Yummly includes weak ingredients lists and the title of the dish, but it does not include food labels.

\textbf{Automatic similarity kernels.} SNaCK is not specific to any specific kernel representation, so we compare two kinds of similarity measures. \emph{Food Kernel 1} is a semantic similarity measure of the best matching between two foods' ingredient lists, and \emph{Food Kernel 2} is a visual similarity measure based on a convolutional neural network. To create $K^{\text{word2vec}}_{i,j}$ \emph{(Food Kernel 1)}, let $I_i$ and $I_j$ be food $i$ and $j$'s ingredients lists from Yummly. Let $w(\cdot)$ be an ingredient's \emph{word2vec}\cite{word2vec} representation, scaled to unit norm, and let cost matrix $C(a,b) = w(a) \cdot w(b)$ for $a \in I_i, b \in I_j$. Finally, let $f : I_i \to I_j$ be the maximum-weight assignment between the two ingredient lists. Then, $K^{\text{word2vec}}_{i,j} = - \sum_{a \in I_i} C(a, f(a))$. This way, \emph{Food Kernel 1} determines foods that share many common ingredients are more similar than foods that have many dissimilar ingredients.

To build \emph{Food Kernel 2}, we fine-tuned a CNN to predict a food label. Because \emph{Yummly-10k} does not have any labels, we train on the \emph{Food-101} dataset from~\cite{bossard_food-101mining_2014}. Similarly to our earlier bird experiments, our network is a variation of the ``Network-in-Network'' model trained to classify 101 different foods. It was trained for 20,000 iterations on an Amazon EC2 GPU instance by replacing the last layer and reducing the learning rate. The final kernel is defined as the Euclidean distance between these CNN features. Our CNN model provides an excellent kernel to start from: when trained via the standard Food-101 protocol, this model achieves rank 1 classification accuracy of 73.5\%. The previous best accuracy on this dataset is 56.40\% from~\cite{bossard_food-101mining_2014}; the best non-CNN is 50.76\%. Of course, building a good classification model is not our focus, but we report this accuracy to show that the automatic kernel we use is effective at distinguishing different foods.

\begin{figure}[t]
  \includegraphics[width=0.9\linewidth]{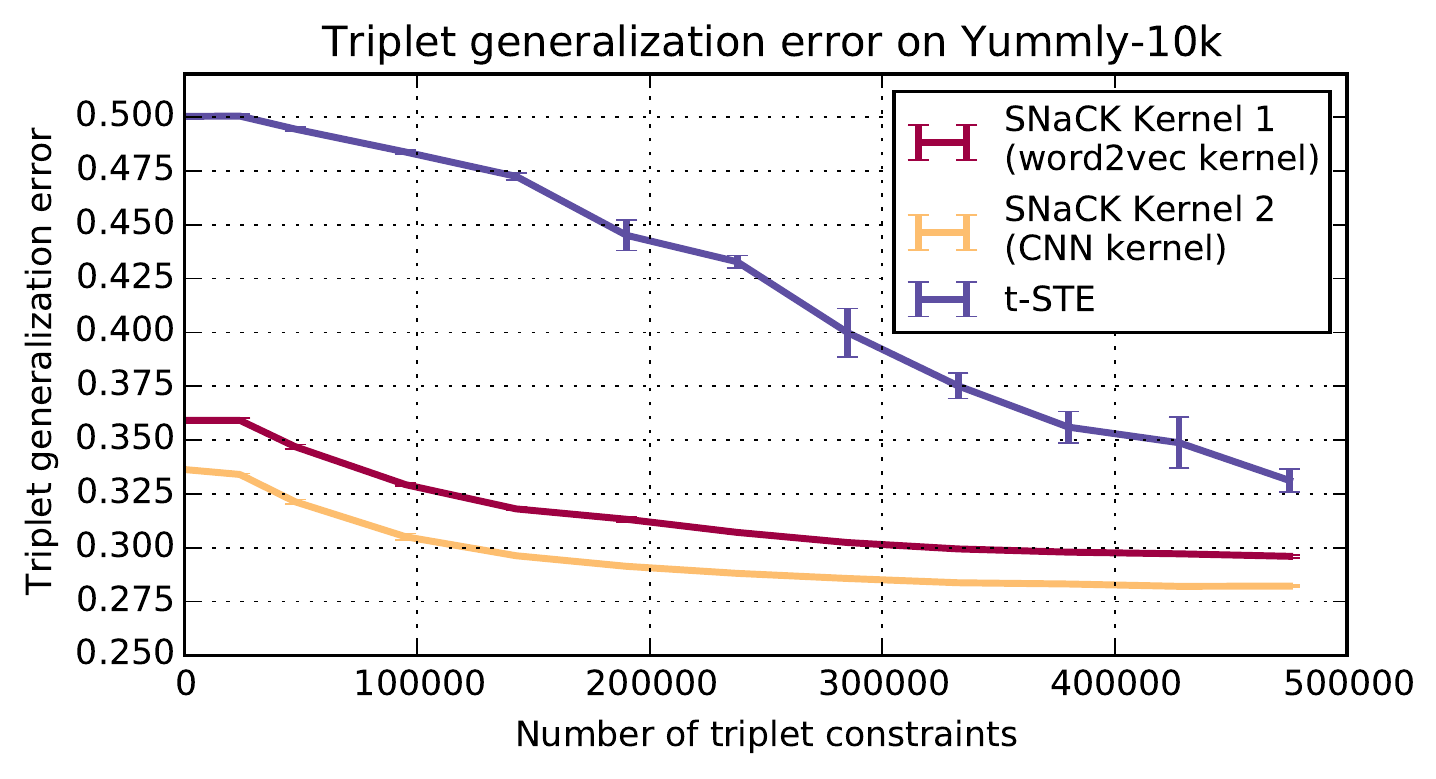}
  \caption{Increasing the number of crowdsourced triplet constraints allows all methods to improve the embedding quality, measured as the fraction of unsatisfied held-out triplet constraints (``triplet generalization error''). However, SNaCK-based methods converge much more quickly than t-STE and require less expert annotation to get a better result. \label{fig:yummly10k-convergence}}
\end{figure}

\textbf{Expert annotation.} Because we want our embedding to properly capture the concept of food taste, we collect our expert annotations directly from humans on Amazon Mechanical Turk using the crowdsourcing interface of~\cite{cost-effective-hits}. For each screen, we show a reference food image $i$ and a grid of 12 food images. The human is asked to ``Please select 4 food images that \emph{taste similar} to the reference food $i$.'' We then generate all possible triplet constraints $\{(i,j,k), j \in S, k \notin S\}$, where $S$ is the user's selection. Each HIT has 10 screens and yields 320 triplet constraints. In total, we collected 958,400 triplet constraints.\footnote{Triplets are available from the companion website, \url{http://vision.cornell.edu/se3/projects/concept-embeddings}}

\textbf{Experiment design.} There are no labels associated with taste in our Yummly data, so we must use other metrics to evaluate the quality of our perceptual embeddings. To do this, we adopt the ``Triplet Generalization Error'' metric common to previous work~\cite{heim_efficient_2015,zhang_jointly_2015,van_der_maaten_stochastic_2012,cost-effective-hits}. We split all triplet constraints into training and testing sets and generate embeddings with varying numbers of training triplet constraints. Triplet generalization error is defined as the fraction of violated testing triplet constraints, which measures the embedding's ability to generalize to constraints the expert did not specify. We compare our two SNaCK kernels to t-STE.

\textbf{Results} are shown in Fig.~\ref{fig:yummly10k-convergence} and an example embedding is shown in Fig.~\ref{fig:food-embeddings}. As more triplet constraint annotations become available, all methods produce embeddings of higher quality. SNaCK with Kernel 2 eventually converges to 28\% while t-STE reaches 33\% error.  Note that t-STE starts from random chance (50\%) because it starts with no information, while SNaCK-based methods initially start with lower error because the Stochastic Neighbor loss on the automatic kernel encourages an initial embedding that contains some fine-grained information. Kernel 2 consistently outperforms Kernel 1, indicating that in this experiment, deep-learned visual features may be a better indication of food taste than the similarity of food ingredient lists. However, even the ``weaker'' semantic ingredient information provides a much better initial kernel than nothing at all.





\subsection{Interactively discovering the structure of pictographic character symbols}\label{sec:experiments_emoji}
\begin{figure}[t]
  \includegraphics[width=\linewidth]{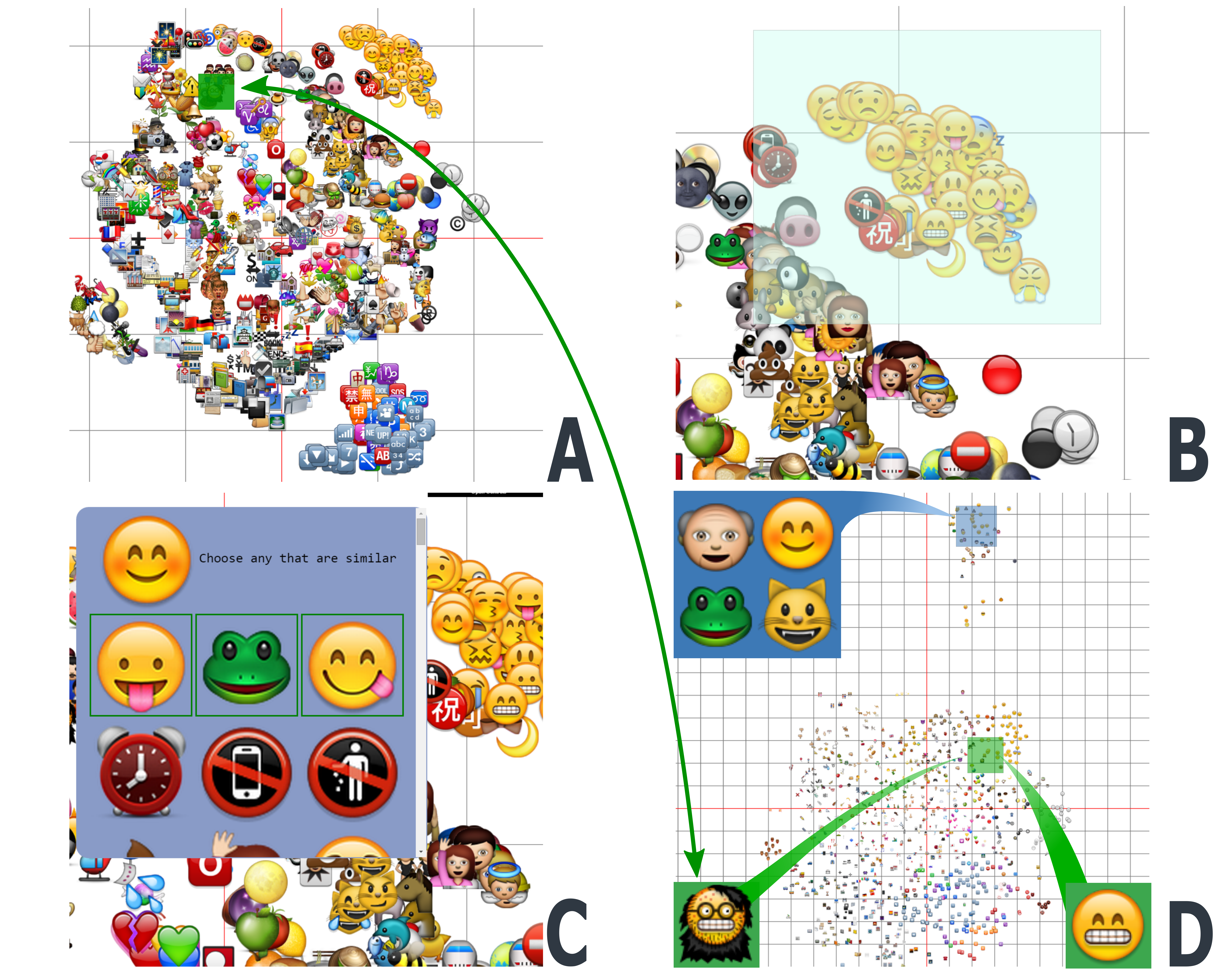}
  \caption{An example GUI used to interactively explore and refine concept embeddings. (A) shows a t-SNE embedding of Emoji using pre-trained ImageNet features. The user selects a set of images (B) and indicates which ones share the same emotion (C). For example, the user selected the smiling frog because it has a similar emotion to the top left image. The updated SNaCK embedding (D) moves smiling emoji away from unrelated images, regardless of the artistic style of the faces. Additionally one of the highlighted fearful faces, separate from the main cluster of faces in (A) has moved to be near faces with a similar expression without collecting triplets between them.}
  \label{fig:emoji}
\end{figure}

In this section we describe possible tools for exploring unlabeled data. We chose to analyze a set of 887 pictographic characters, colloquially known as Emoji. Using CNN features pre-trained on ImageNet, we can create an embedding that does a good job of grouping visually similar Emoji together. However, if the goal is to capture the concept of emotion within the set of Emoji, then similarity of visual features alone may be inadequate. For example, in Fig.~\ref{fig:emoji}.A, a group of yellow faces are clustered at the upper right, but this group contains different emotions and does not contain similar images in other artistic styles.

To interactively refine the embedding, the expert selects a reference Emoji and drags a box around several images. The expert then indicates which of these images share the same emotion as the reference. In the example in Fig.~\ref{fig:emoji}, a smiling Emoji was selected and compared to all the Emoji in the green box (Fig.~\ref{fig:emoji}.B). After two bounding box selections and a few minutes of work, we are able to collect 20,000 triplets and separate many of the smiling Emoji from the rest of the embedding.  From here, we could further inspect these Emoji and separate the emotion of laughing from smiling.

As mentioned in the MNIST experiments, the SNaCK embeddings are capable of taking advantage of visual cues when triplet information is not available. An example of this can be seen in Fig.~\ref{fig:emoji}.D. A fearful face with glasses is moved from the left side of embedding to be near other faces with similar expressions. SNaCK was able to do this without requiring triplets to be collected between these faces. These examples give a brief illustration of how SNaCK can be useful for examining unlabeled data.
\section{Conclusion}
Our SNaCK algorithm can learn concept embeddings by combining human expertise with machine similarity. We showed that SNaCK can help experts quickly label new sets of woodpeckers and vireos, build training sets for classifiers in a semi-supervised fashion, and capture the perceptual structure of food taste. We also presented a snapshot of a tool that can help experts interactively explore and refine a set of pictographic characters. In the future, we will pursue intelligent sampling for active learning of embeddings, and will extend our system to explore large video datasets.
\section{Acknowledgments}
This work was partially supported by an NSF Graduate Research Fellowship award (NSF DGE-1144153, Author 1), a Google Focused Research award (Author 4), and AOL-Program for Connected Experiences (Authors 1 and 4). We also wish to thank Laurens van der Maaten and Andreas Veit for insightful discussions.



{\small
\bibliographystyle{ieee}
\bibliography{main}

\begin{thebibliography}{10}\itemsep=-1pt

\bibitem{abu-mostafa_machines_1995}
Y.~S. Abu-Mostafa.
\newblock Machines that learn from hints.
\newblock {\em Scientific American}, 272:64--69, 1995.

\bibitem{amid_multiview_2015}
E.~Amid and A.~Ukkonen.
\newblock Multiview {Triplet} {Embedding}: {Learning} {Attributes} in
  {Multiple} {Maps}.
\newblock In {\em Proceedings of the 32nd {International} {Conference} on
  {Machine} {Learning} ({ICML}-15)}, pages 1472--1480, 2015.

\bibitem{beijbom2012automated}
O.~Beijbom, P.~J. Edmunds, D.~I. Kline, B.~G. Mitchell, and D.~Kriegman.
\newblock Automated annotation of coral reef survey images.
\newblock In {\em CVPR}. IEEE, 2012.

\bibitem{label-prop}
Y.~Bengio, O.~Delalleau, and N.~L. Roux.
\newblock {\em Label Propagation and Quadratic Criterion}, pages 193--216.
\newblock MIT Press, 2006.

\bibitem{bilenko_integrating_2004}
M.~Bilenko, S.~Basu, and R.~J. Mooney.
\newblock Integrating constraints and metric learning in semi-supervised
  clustering.
\newblock In {\em ICML}. ACM, 2004.

\bibitem{biswas_simultaneous_2013}
A.~Biswas and D.~Parikh.
\newblock Simultaneous {Active} {Learning} of {Classifiers} and {Attributes}
  via {Relative} {Feedback}.
\newblock In {\em CVPR}, June 2013.

\bibitem{bossard_food-101mining_2014}
L.~Bossard, M.~Guillaumin, and L.~Van~Gool.
\newblock Food-101--{Mining} {Discriminative} {Components} with {Random}
  {Forests}.
\newblock In {\em ECCV}. Springer, 2014.

\bibitem{BransonVBP14}
S.~Branson, G.~V. Horn, S.~Belongie, and P.~Perona.
\newblock Bird species categorization using pose normalized deep convolutional
  nets.
\newblock In {\em BMVC}, Nottingham, 2014.

\bibitem{branson_ignorant_2014}
S.~Branson, G.~V. Horn, C.~Wah, P.~Perona, and S.~Belongie.
\newblock The {Ignorant} {Led} by the {Blind}: {A} {Hybrid} {Human}-{Machine}
  {Vision} {System} for {Fine}-{Grained} {Categorization}.
\newblock {\em IJCV}, 108(1-2):3--29, Feb. 2014.

\bibitem{demiralp_learning_2014}
C.~Demiralp, M.~Bernstein, and J.~Heer.
\newblock Learning {Perceptual} {Kernels} for {Visualization} {Design}.
\newblock {\em IEEE Trans. on Visualization and Computer Graphics},
  20(12):1933--1942, Dec. 2014.

\bibitem{demiralp_visual_2014}
C.~Demiralp, C.~Scheidegger, G.~Kindlmann, D.~Laidlaw, and J.~Heer.
\newblock Visual {Embedding}: {A} {Model} for {Visualization}.
\newblock {\em IEEE Computer Graphics and Applications}, 34(1):10--15, Jan.
  2014.

\bibitem{farrell_birdlets:_2011}
R.~Farrell, O.~Oza, N.~Zhang, V.~I. Morariu, T.~Darrell, and L.~S. Davis.
\newblock Birdlets: {Subordinate} categorization using volumetric primitives
  and pose-normalized appearance.
\newblock In {\em ICCV}, 2011.

\bibitem{gomes_crowdclustering_2011}
R.~Gomes, {Welinder, Peter}, {Krause, Andreas}, and {Perona, Pietro}.
\newblock Crowdclustering.
\newblock {\em NIPS}, 2011.

\bibitem{hadsell_dimensionality_2006}
R.~Hadsell, S.~Chopra, and Y.~LeCun.
\newblock Dimensionality reduction by learning an invariant mapping.
\newblock In {\em CVPR}, 2006.

\bibitem{heim_efficient_2015}
E.~Heim, M.~Berger, L.~M. Seversky, and M.~Hauskrecht.
\newblock Efficient {Online} {Relative} {Comparison} {Kernel} {Learning}.
\newblock {\em arXiv preprint arXiv:1501.01242}, 2015.

\bibitem{jia2014caffe}
Y.~Jia, E.~Shelhamer, J.~Donahue, S.~Karayev, J.~Long, R.~Girshick,
  S.~Guadarrama, and T.~Darrell.
\newblock Caffe: Convolutional architecture for fast feature embedding.
\newblock {\em arXiv preprint arXiv:1408.5093}, 2014.

\bibitem{kendall1948rank}
M.~G. Kendall.
\newblock {\em Rank correlation methods.}
\newblock Griffin, 1948.

\bibitem{kleindessner_uniqueness_2014}
M.~Kleindessner and U.~von Luxburg.
\newblock Uniqueness of {Ordinal} {Embedding}.
\newblock {\em JMLR}, 2014.

\bibitem{lad_interactively_2014}
S.~Lad and D.~Parikh.
\newblock Interactively guiding semi-supervised clustering via attribute-based
  explanations.
\newblock In {\em ECCV}. Springer, 2014.

\bibitem{lee_learning_2011}
Y.~J. Lee and K.~Grauman.
\newblock Learning the easy things first: {Self}-paced visual category
  discovery.
\newblock In {\em CVPR}, pages 1721--1728, June 2011.

\bibitem{lin_network_2013}
M.~Lin, Q.~Chen, and S.~Yan.
\newblock Network {In} {Network}.
\newblock {\em arXiv:1312.4400 [cs]}, Dec. 2013.
\newblock arXiv: 1312.4400.

\bibitem{lu_constrained_2008}
Z.~Lu and M.~Carreira-Perpinan.
\newblock Constrained spectral clustering through affinity propagation.
\newblock In {\em CVPR}, June 2008.

\bibitem{mcfee}
B.~{McFee}.
\newblock {\em More like this: machine learning approaches to music
  similarity}.
\newblock PhD thesis, University of California, San Diego, May 2012.

\bibitem{word2vec}
T.~Mikolov, K.~Chen, G.~Corrado, and J.~Dean.
\newblock Efficient estimation of word representations in vector space.
\newblock {\em arXiv preprint arXiv:1301.3781}, 2013.

\bibitem{miller1956magical}
G.~A. Miller.
\newblock The magical number seven, plus or minus two: some limits on our
  capacity for processing information.
\newblock {\em Psychological review}, 63(2):81, 1956.

\bibitem{DBLP:journals/corr/RussakovskyDSKSMHKKBBF14}
O.~Russakovsky, J.~Deng, H.~Su, J.~Krause, S.~Satheesh, S.~Ma, Z.~Huang,
  A.~Karpathy, A.~Khosla, M.~S. Bernstein, A.~C. Berg, and L.~Fei{-}Fei.
\newblock Imagenet large scale visual recognition challenge.
\newblock {\em CoRR}, abs/1409.0575, 2014.

\bibitem{schroff_facenet:_2015}
F.~Schroff, D.~Kalenichenko, and J.~Philbin.
\newblock {FaceNet}: {A} {Unified} {Embedding} for {Face} {Recognition} and
  {Clustering}.
\newblock {\em arXiv:1503.03832 [cs]}, Mar. 2015.
\newblock arXiv: 1503.03832.

\bibitem{googlenet}
C.~Szegedy, W.~Liu, Y.~Jia, P.~Sermanet, S.~Reed, D.~Anguelov, D.~Erhan,
  V.~Vanhoucke, and A.~Rabinovich.
\newblock Going {Deeper} with {Convolutions}.
\newblock {\em arXiv:1409.4842 [cs]}, Sept. 2014.
\newblock arXiv: 1409.4842.

\bibitem{tamuz_adaptively_2011}
O.~Tamuz, C.~Liu, S.~Belongie, O.~Shamir, and A.~T. Kalai.
\newblock Adaptively {Learning} the {Crowd} {Kernel}.
\newblock In {\em {ICML}}, 2011.

\bibitem{tang_enhancing_2007}
W.~Tang, H.~Xiong, S.~Zhong, and J.~Wu.
\newblock Enhancing {Semi}-supervised {Clustering}: {A} {Feature} {Projection}
  {Perspective}.
\newblock In {\em SIGKDD}, 2007.

\bibitem{van_der_maaten_visualizing_2008}
L.~Van~der Maaten and G.~Hinton.
\newblock Visualizing data using t-{SNE}.
\newblock {\em JMLR}, 9(2579-2605):85, 2008.

\bibitem{van_der_maaten_stochastic_2012}
L.~Van~der Maaten and K.~Weinberger.
\newblock Stochastic triplet embedding.
\newblock In {\em {IEEE} {Int.} {Workshop} on {Machine} {Learning} for {Signal}
  {Processing}}, 2012.

\bibitem{WahCUB_200_2011}
C.~Wah, S.~Branson, P.~Welinder, P.~Perona, and S.~Belongie.
\newblock {The Caltech-UCSD Birds-200-2011 Dataset}.
\newblock Technical Report CNS-TR-2011-001, California Institute of Technology,
  2011.

\bibitem{wah_similarity_2014}
C.~Wah, G.~V. Horn, S.~Branson, S.~Maji, P.~Perona, and S.~Belongie.
\newblock Similarity {Comparisons} for {Interactive} {Fine}-{Grained}
  {Categorization}.
\newblock {\em CVPR}, 2014.

\bibitem{wang_learning_2014}
J.~Wang, Y.~song, T.~Leung, C.~Rosenberg, J.~Wang, J.~Philbin, B.~Chen, and
  Y.~Wu.
\newblock Learning {Fine}-grained {Image} {Similarity} with {Deep} {Ranking}.
\newblock {\em arXiv:1404.4661 [cs]}, Apr. 2014.
\newblock arXiv: 1404.4661.

\bibitem{wang_constrained_2012}
X.~Wang, B.~Qian, and I.~Davidson.
\newblock On constrained spectral clustering and its applications.
\newblock {\em Data Mining and Knowledge Discovery}, 28(1):1--30, Sept. 2012.

\bibitem{cost-effective-hits}
M.~J. Wilber, I.~S. Kwak, and S.~J. Belongie.
\newblock Cost-effective hits for relative similarity comparisons.
\newblock In {\em AAAI Conference on Human Computation and Crowdsourcing},
  2014.

\bibitem{wu_online_2013}
P.~Wu, S.~C. Hoi, H.~Xia, P.~Zhao, D.~Wang, and C.~Miao.
\newblock Online {Multimodal} {Deep} {Similarity} {Learning} with {Application}
  to {Image} {Retrieval}.
\newblock In {\em {ACM} {International} {Conference} on {Multimedia}}, 2013.

\bibitem{xing_distance_2002}
E.~P. Xing, M.~I. Jordan, S.~Russell, and A.~Ng.
\newblock Distance metric learning with application to clustering with
  side-information.
\newblock In {\em NIPS}, 2002.

\bibitem{yi_semi-crowdsourced_2012}
J.~Yi, R.~Jin, S.~Jain, T.~Yang, and A.~K. Jain.
\newblock Semi-crowdsourced clustering: {Generalizing} crowd labeling by robust
  distance metric learning.
\newblock In {\em NIPS}, 2012.

\bibitem{yi_semi-supervised_2013}
J.~Yi, L.~Zhang, R.~Jin, Q.~Qian, and A.~Jain.
\newblock Semi-supervised {Clustering} by {Input} {Pattern} {Assisted}
  {Pairwise} {Similarity} {Matrix} {Completion}.
\newblock In {\em ICML}, 2013.

\bibitem{zhang_jointly_2015}
L.~Zhang, S.~Maji, and R.~Tomioka.
\newblock Jointly {Learning} {Multiple} {Perceptual} {Similarities}.
\newblock {\em arXiv:1503.01521 [cs, stat]}, Mar. 2015.
\newblock arXiv: 1503.01521.

\end{thebibliography}
}

\end{document}